# How human judgment impairs automated deception detection performance

Bennett Kleinberg[1,2] & Bruno Verschuere[2]

[1]Department of Security and Crime Science, University College London
[2]Department of Psychology, University of Amsterdam

**ABSTRACT:**

Background: Deception detection is a prevalent problem for security practitioners. With a need for more large-scale approaches, automated methods using machine learning have gained traction. However, detection performance still implies considerable error rates. Findings from other domains suggest that hybrid human-machine integrations could offer a viable path in deception detection tasks.

Method: We collected a corpus of truthful and deceptive answers about participants' autobiographical intentions (n=1640) and tested whether a combination of supervised machine learning and human judgment could improve deception detection accuracy. Human judges were presented with the outcome of the automated credibility judgment of truthful and deceptive statements. They could either fully overrule it (hybrid-overrule condition) or adjust it within a given boundary (hybrid-adjust condition).

Results: The data suggest that in neither of the hybrid conditions did the human judgment add a meaningful contribution. Machine learning in isolation identified truth-tellers and liars with an overall accuracy of 69%. Human involvement through hybrid-overrule decisions brought the accuracy back to the chance level. The hybrid-adjust condition did not deception detection performance. The decision-making strategies of humans suggest that the truth bias - the tendency to assume the other is telling the truth - could explain the detrimental effect.

Conclusion: The current study does not support the notion that humans can meaningfully add to the deception detection performance of a machine learning system.

## INTRODUCTION

Determining who is lying and who is telling the truth is at the core of the legal system and has sparked the interest of the academic community for decades. While some approaches rely on physiological measurements such as brain potential or skin conductance (for a recent overview see Rosenfeld, 2018), others look at the verbal content (Oberlader et al., 2016) and the linguistic properties of statements made by liars and truth-tellers (e.g., Pérez-Rosas & Mihalcea, 2014). Until recently, the majority of approaches has focused on the detection of lies about past events such as a classic crime scenario where investigators try to establish who committed a crime. Since a few years, however, the academic research on deception detection has moved closer to the practitioners' needs of being able to assess whether someone might be a threat and might hold malicious intent. Such an approach is proactive and in line with the crime prevention task of law enforcement.

In addition to the focus on prevention security practitioners operate increasingly in large-scale contexts. For example, border control settings or airport security control require the screening of vast amounts of people. These contexts require approaches that are structurally different from those applied in murder investigations, for example (for a review on needs for large-scale deception detection methods, see Kleinberg et al., 2019). The principal concern with approaches that require extensive human involvement is that these are hard to scale up – both in terms of engaging with examinees and in deciding about the credibility of a statement. Among the promising candidates for large-scale purposes is the use of information provided by a person. Meta-analytical research agrees that the verbal (or linguistic) approach to deception detection is significantly better than the chance level (Hauch et al., 2017; Oberlader et al., 2016; Vrij et al., 2017). Classical deception detection typically requires 1-on-1 interaction in an interview and human involvement in scoring the verbal transcripts of the interviews. Research on computational efforts of understanding human language has shown that methods from natural language processing can be used to analyze the verbal content automatically and learn to estimate the credibility of a statement (Kleinberg et al., 2018; Mihalcea & Strapparava, 2009; Ott et al., 2011; Pérez-Rosas et al., 2015, 2017).

One fundamental problem of deception research is that the accuracy of correctly identifying liars and truth-tellers on average exceeds the guessing level only by about 20 accuracy points (see Kleinberg et al., 2019a; Oberlader et al., 2016) both for manual coding procedures (Vrij et al., 2017) as well as for fully automated deception detection (e.g., Mihalcea & Strapparava, 2009; Pérez-Rosas & Mihalcea, 2014; Soldner et al., 2019). In particular, for large-scale settings where the base rate of persons of interest is often low, these accuracy rates are not satisfactorily (see Honts & Hartwig, 2014; Kleinberg et al., 2019). In this paper, we test whether deception detection performance can be augmented by combining two distinct modes of decision-making: automated classification and human judgment.

**Automated versus human deception detection**

Both fully automated and manual, human approaches to deception detection follow the same goal. However, how they arrive at a truth vs lie judgment are structurally different. Human deception detection includes a human judge who reads a statement or watches a video that contains truthful and deceptive accounts and is then asked to decide whether they believe the person or not. Research agrees that this task is difficult and several studies and meta-analyses have pointed out that humans perform at the chance level (Bond & DePaulo, 2006; Hartwig et al., 2017; Hartwig & Bond, 2011). A possible explanation for guessing-level performance is that humans tend to believe rather than suspect deception. That truth bias leads to typically higher accuracy in truth-detection compared to lie detection (Levine, 2014, 2018). Another explanation is that humans typically hold incorrect beliefs about cues to deception. Several studies support the idea that humans rely on unsupported cues (Hartwig & Bond, 2011) regardless of their expertise or training (Bogaard et al., 2016; Hauch et al., 2016).

In contrast, automated deception detection works without human involvement and is typically done on the verbal transcript of a spoken statement (Pérez-Rosas et al., 2015) or written texts directly (Kleinberg et al., 2018; Pérez-Rosas & Mihalcea, 2014). Using the text as the data, one extracts features from the text (e.g., in the form of word frequencies or psycholinguistic variables) and then utilizes supervised machine learning to predict the outcome label (deceptive vs truthful) of texts. Commonly used features are the variables derived from the Linguistic Inquiry and Word Count software (LIWC, Pennebaker et al., 2015) and frequency count approaches using word or sequence occurrences ($n$-grams), part-of-speech tags (e.g., nouns, verbs, adjectives, Ott et al., 2011) or named entities (Kleinberg et al., 2017). While the classification algorithm (e.g., support vectors, random forests, Naïve Bayes) and its specifications can differ, the objective is always to find combinations of features that best classify training examples. Underlying each binary classification are class probabilities which indicate the certainty of the machine judgment (e.g. 0.60 is less certain than 0.99 for given class membership, for an overview of machine learning for behavioural science, see Yarkoni & Westfall, 2017).

Currently, the deception detection performance of the automated approach ranges between 64% and 80% (see Fornaciari & Poesio, 2013; Kleinberg et al., 2018; Mihalcea & Strapparava, 2009; Pérez-Rosas et al., 2015; Pérez-Rosas & Mihalcea, 2014).

To date, both modes of decision-making (human and automated) were used in isolation. However, triaging systems and human-machine integration were shown to be successful in related detection problems. It is mainly unexamined for deception detection in general, and the detection of deceptive intentions in specific, whether the integration of these two decision-making modes into hybrid approaches is beneficial.

**Hybrid approaches**

Approaches that integrate machine and human judgment are understudied in deception detection but are commonly used in online content moderation (Jhaver et al., 2019; van der Vegt et al., 2019) and gain traction in medical diagnoses (Bulten et al., 2020). The workflow usually starts with a decision made by a machine learning system of which a portion of cases is then forwarded to humans for their manual review. In online content removal tasks, only uncertain cases might be forwarded to human reviewers, whereas in medical cases the final review of all cases lies with a medical practitioner. Both cases share a characteristic with current deception detection, namely that of large-scale and low base rate problems. The underlying rationale of hybrid approaches is that automated judgments can aide the human decision-maker yielding overall better performance than either mode in isolation.

In the case of verbal deception detection, that promise translates to the dilemma between vast amounts of information and making sense of contextual pieces of information. While machine learning allows classifying high-dimensional data, it currently lacks the means to quantify and hence measure concepts that are semantically heavy such as the plausibility of information in a specific context.[1] The latter, however, comes relatively easy for humans who read a statement. For example, suppose a person provides information about their upcoming flight to London. The statement might include information about the attractions to visit on a day. For a human, it might immediately flag as strange or suspicious if the person stated that they would take the tube from Stansted Airport (since the London tube network does not extend to that place). Someone who intended to visit London would probably not have provided such inaccurate information. Automated systems struggle to extract the implausibility and falsehood of such a statement. To that end, human judges could help since they can interpret context but lack the cognitive capacity to make inferences from high-dimensional data. Hybrid approaches could, therefore, be a means to utilize the advantages of both modes: the capacity to process and make decisions based on vast and complex data as well as the ability to spot contextual inconsistencies and implausible information.

To date, only limited research is available on a hybrid deception detection approach. Of the two studies available, one does not detail the decision-support system of human-machine collaboration (Quijano-Sánchez et al., 2018). The other work used deceptive and genuine hotel reviews first assessed by a supervised machine learning classification utilizing linguistic variables of the reviews (Harris, 2019). The label predicted by the classification system was then presented to human judges along with the original review, the value of the review on each of the LIWC variables, as well as the average LIWC score across all reviews. Human assessors were then asked to determine whether they think the review is fake or not. The classification accuracy of the hybrid approach was virtually the same as that of the best machine learning model (95.1% vs 94.9%) and some methodological problems persist.

---

[1] Although vector space models using word embeddings start to tap into semantic relationships of words, these are currently not yet able to grasp the plausibility of claims and contextual information.

First, a potential explanation of the high accuracies could be that the models pick up structural differences between fake and genuine reviews that are unrelated to the deception. Typically, genuine reviews are written by people who visited a hotel while fake reviews are fabricated by crowdsource workers who have never been to the hotel they write about. Therefore, the linguistic differences could be a function of knowledge of the property rather than deception. Compared to the typical deception detection accuracy, hotel reviews are a stark outlier and might hence create a ceiling effect for detection performance. Hotel review detection is typically an easier task for machine learning approaches as well as human assessors than deception detection on actual events or planned activities (Ott et al., 2011, 2013)[2]. The potential ceiling effect here could have hindered improvements through the hybrid method. Second, humans were forced to make a binary decision and could hence not incorporate any uncertainty in their judgment. Neither did the presentation of the machine model results did not consider the uncertainty either (i.e., human assessors could not tell whether the decision was a boundary case or not). The usefulness of combining human and machine efforts for deception detection about future events in an experimental setting is unclear but could potentially offer a solution to augment the decision-making process.

**Aims of this paper**

This paper aims to examine how computer-automated deception detection can be combined with human judgment in a setting of deceptive intentions. Specifically, we investigate whether human judges can adjust the class probabilities of supervised learning to allow for better classification performance. We include a human baseline and two hybrid conditions. One hybrid condition allows the human to fully overrule the machine judgment while the other constraint the allowed deviation from the machine judgment. This study is the first to explore how both modes of decision-making can be combined to detect deceptive intentions and looks at how human judges engage with machine judgment.

## METHOD

This study was approved by the local IRB.

**Transparency statement**

The data collected for the study reported here are publicly available on the Open Science Framework at https://osf.io/45z7e/?view_only=79a4f6cc1f9f40fe8d68ec405c9dd387

---

[2] A potential explanation of the high accuracies could be that the models pick up structural differences between fake and genuine reviews that are unrelated to the deception. Typically, genuine reviews are written by people who visited a hotel while fake reviews are fabricated by crowdsource workers who have never been to the hotel they write about. Therefore, the linguistic differences could be a function of knowledge of the property rather than deception.

**Corpus of truthful and deceptive statements**

We used a web-interface where participants were asked to provide a statement about their most significant non-work-related activity in the next seven days.[3] All participants were batch-wise[4] allocated to the truthful or deceptive condition, and the data were collected through the crowdsourcing platform Prolific Academic. Upon entering their activity (e.g. "attending my brother's wedding"), the participants were informed that their statements were later read by human experts and assessed by an automated system. Their task was to provide an as convincing as possible answer to two brief questions (Q1: "Please describe your activity as specific as possible", Q2: "Which information can you give us to reassure us that you are telling the truth").

Those in the truthful condition were asked to answer these questions truthfully. Participants in the deceptive condition were allocated to someone else's activity. They were presented three activities from participants in the truthful condition and asked to indicate which ones did not apply to them. From these activities, we randomly selected one and instructed the participant to pretend that this would be their most important activity for the next week. They then received the same instructions to be as convincing as possible when answering the two questions.

At the end of the task, the participants indicated how motivated they were to appear convincing (from 0 = not motivated at all, to 10 = very motivated), how certain it was that they would carry out their actual activity (from 0 = not certain at all, to 10 = absolutely certain), and what their initial task instructions were.

2,027 participants provided a statement (see Appendix 1 for age and gender information). We excluded those who (i) failed the manipulation check ($n$=29), (ii) provided no or too short input[5] ($n$=345), and (iii) whose answers to the second question resembled their answer on the first question too much[6] ($n$=13). Overall, the participants in the final sample reported high motivation ($M$=8.45, $SD$=1.58). The final corpus consisted of 1,640 statements with two answers each and corpus lengths of 87,555 words (Q1) and 65,948 words (Q2, see Table 1).

Table 1. *Corpus descriptive statistics.*

|  | Q1: Please describe your activity as specific as possible | | | Q2: Which information can you give us to reassure us that you are telling the truth | | |
|---|---|---|---|---|---|---|
|  | M (SD) | Median | Range | M (SD) | Median | Range |
| Number of words | 53.39 (32.20) | 46 | 15; 274 | 40.21 (26.45) | 34 | 10; 308 |
| Number of sentences | 2.59 (1.61) | 2 | 1; 13 | 2.08 (1.31) | 2 | 1; 12 |
| Characters per word | 4.73 (0.35) | 4.70 | 3.43; 6.75 | 4.74 (0.42) | 4.71 | 3.07; 6.64 |

---

[3] The activity should be "specific, have a clear start and an end time, and it should not be a continuous or daily activity".

[4]. The batch-wise allocation ensured that we could match the autobiographical activities from the participants in the truthful condition to those in the deceptive condition.

[5] Too short is defined here as an answer shorter than 15 words.

[6] We used a string similarity of 0.40 as a criterion. If the characters overlapped by more 0.40, we excluded the participant.

**Machine learning classification**

We used supervised machine learning to classify truthful and deceptive answers. We extracted the following features from the responses and reported the classification metrics for each.

Linguistic Inquiry and Word Count (LIWC) variables: we used all 93 categories of the LIWC as a feature set. The LIWC aims to measure linguistic and psycholinguistic processes through a word count lexicon approach (see Fornaciari & Poesio, 2013; Kleinberg et al., 2018; Pérez-Rosas & Mihalcea, 2014).

Relative part-of-speech (POS) frequencies: we extracted the POS of each word and calculated the frequency of each relative to the overall number of words. The POS tags were extracted according to the Universal Dependencies scheme (https://universaldependencies.org/u/pos/).

For the classification exercises, we used 80% of the data ($n$=1313) for training and tested the final algorithm on the held-out 20% ($n$=327). On the training set, we used 10-fold cross-validation with ten repetitions and utilized a vanilla random forest as the learning algorithm.

**Human judgment and hybrid approach**

*Experimental task*

We built a web-app for the judgment of the truthful and deceptive statements. The answers to the first question ("Please describe your activity as specific as possible") were used from the 325[7] data points of the hold-out test set from the automated classification. That ensured that the same texts were rated by both an independent automated classification and human judges.

We instructed the participants to read the original answer and to make a judgment about its veracity. The participants were informed that the answers stemmed from a previous experiment where the participants were instructed to either answer truthfully or fabricate their answer about their most important upcoming activity. Each participant was randomly allocated to one of three conditions.

In the **human baseline condition**, we asked the participant to read the statement and indicate their judgment on a slider from 0 (=certainty truthful) to 100 (=certainly deceptive, Table 2). The range was intended to mimic the class probabilities of the machine learning classification and to allow for a quantification of the participants' uncertainty. The slider for each judgment started at the neutral midpoint of 50, and the participants could freely adjust it.

The **hybrid-overrule condition** differed in the slider starting position. Participants were told that the starting point reflected the judgment of an "automated artificial intelligence program" that evaluated the texts. We used the class probabilities of the best automated model using the LIWC features (see Table 2). Their task was to adjust that judgment to their best knowledge.

The **hybrid-adjust condition** differed from the hybrid-overrule condition in that the participants could not use the full slider range but were constrained by adjusting the slider starting position only

---

[7] We used 325 instead of the full 327 to have an even number as the required sample of judges for three judgments per condition and five judgments per participant.

10 points to either direction. These boundaries were indicated as a green area around the starting position.

All participants received two instruction trials with judgments according to their condition's constraints (e.g. "move the slider towards a more deceptive judgment") and could proceed once they correctly followed the instructions. In all three conditions, each participant judged five different statements, and we aimed for three judgments per statement and condition in total. Participants received GBP 1.25 for participation in this 5 to 10-minute task (equivalent to GBP 7.50 – 15.00/h). For each correct judgment, they were awarded GBP 0.25 extra, making the total possible reward per participant GBP 2.50.

Table 2. *Instructions and condition specifications for the phase 2 data collection.*

| Condition | Instructions | Slider start position | Condition constraints |
|---|---|---|---|
| Human baseline | Please indicate your judgment as follows:<br>- below each statement you will see a slider with values from 0 (= certainly truthful) to 100 (= certainly deceptive)<br>- use the slider to indicate how truthful or deceptive you think the statement is<br>- values on the left = you judge a statement to be more truthful<br>- values on the right = you judge the statement to be more deceptive<br>- the slider is set by default to a neutral point of 50 (i.e. indecisive between truthful and deceptive)<br>- the starting position of the slider is also indicated by a small black line on the slider<br>- move the slider to the left if you think the statements is more likely to be truthful, move the slider to the right to indicate you think the statements is more likely to be deceptive<br>- the more you move the slider to the extremes, the more certainty you indicate with your judgment (i.e. values closer to the middle suggest that you are less certain of your judgment) | Neutral midpoint 50. | None. |
| Hybrid overrule | Same as the human baseline condition, except:<br>- You will see that the slider is set by default at a specific judgment. This point reflects the judgment of an artificial intelligence (AI) programme that was trained on some statements and then judged the truthfulness of the statements you are about to read<br>- adjust the AI judgment by moving the slider the starting position of the slider is also indicated by a small black line on the slider<br>- you can make use of the full range of values | Class probability of machine learning classification | None. |
| Hybrid adjust | Same as the hybrid overrule condition, except:<br>- you are allowed to adjust the AI judgment up to 10 points to the left or right<br>- the green area shows you the allowed region in which you make a valid judgment | Class probability of machine learning classification | Judgment only allowed to be +/- 10 points of the starting position |

*Participants*

Judgment data of a total of $n$=586 participants were collected (human baseline condition = 35.32%, hybrid-overrule condition = 30.20%, hybrid-adjust condition = 34.47%). Each of the 325 answers was rated on average 3.05 times in each condition. The various judgments (ranging from 0-100) were averaged per statement and condition. There were no exclusion criteria. There were no differences between the three conditions in gender, $X^2(2) = 0.14$, $p = .930$, or age, $F(1, 557) = 1.09$, $p = .298$.

## RESULTS

*Veracity judgment performance metrics*

Table 3 shows the performance metrics for the automated classification and each human condition. For the automated approach, the LIWC feature set resulted in the best accuracy and area under the curve performing significantly above the chance level. In both the human baseline condition and the hybrid-overrule condition, the accuracy did not exceed the chance level. In the hybrid-adjust condition, the performance did exceed the chance level. Note that the adjustments possible for the human judges were here constrained to +/- 10 points. In all three conditions, the true negative rate (=proportion of correctly detected truth-tellers) is higher than the true positive rate.

Table 3. *Performance metrics for all veracity judgment conditions.*

| Judgment condition | Acc. | AUC | TPR | TNR |
|---|---|---|---|---|
| Human baseline$^\$$ | 0.50 [0.45; 0.56] | 0.52 [0.46; 0.58] | 0.24 | 0.78 |
| Hybrid-overrule$^\$$ | 0.51 [0.45; 0.58] | 0.49 [0.43; 0.55] | 0.25 | 0.76 |
| Hybrid-adjust$^\$$ | 0.67 [0.62; 0.72]* | 0.74 [0.68; 0.79] | 0.60 | 0.74 |
| Machine learning: LIWC | 0.69 [0.63; 0.74]* | 0.75 [0.69; 0.80] | 0.76 | 0.60 |
| Machine learning: POS | 0.64 [0.58; 0.69]* | 0.67 [0.61; 0.73] | 0.71 | 0.56 |

Note. Squared brackets denote the 95% confidence interval. TPR = true positive rate (sensitivity), TNR = true negative rate (specificity) with respect to deceptive answers as the positive class. $^\$$ = for the accuracy we used a human rating of 52 (= chance level) as the threshold. Three indecisive judgments that were exactly 52 were excluded. * = sign. better than chance level at p < .001 (random baseline = 0.52)

*Human decision-making strategies*

To understand how the human judges made their decision, we looked at the difference between human judgment and the initial anchor slider starting point. Figures 1 and 2 show how much the human judges adjusted the class probabilities into which direction. There is evidence that in the majority of cases, human judgment tended to adjust the rating towards "more truthful". This is further supported by the high true negative rates in Table 3 (i.e. truthful and deceptive statements were judged as more truthful). That truth tendency was stronger in the hybrid-overrule than in the human baseline and hybrid-adjust condition (Appendix 2). Interestingly, when we restrained the adjustment in the hybrid-adjust condition, the tendency to move deceptive statements towards the truthful end to a higher degree than with truthful statements, disappeared. Thus, the constraints of the condition could have restrained them from making the deceptive statements even more truthful.

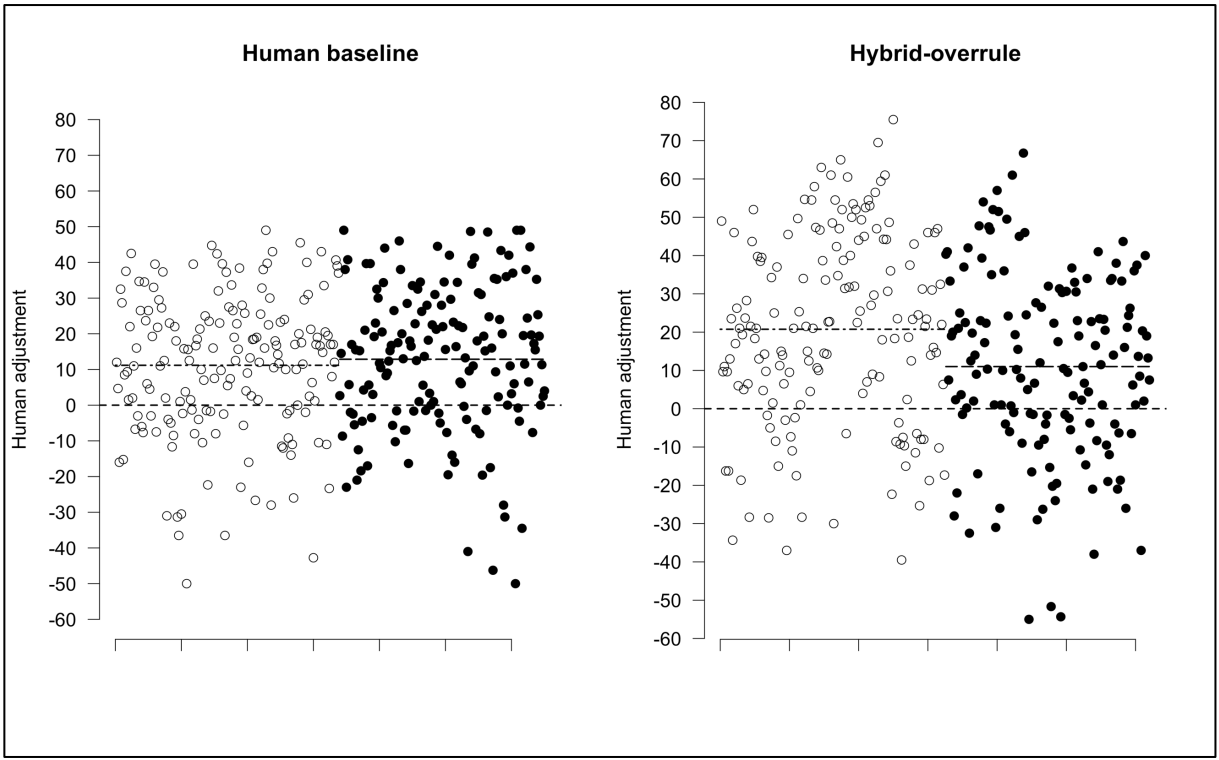

Figure 1. Difference between human judgment and the anchoring position for deceptive (circles) and truthful (dots) answers for the human baseline and hybrid-overrule condition. Positive values indicate that the judgment was adjusted to be more truthful. Negative values indicate that the judgment was adjusted to the deceptive side.

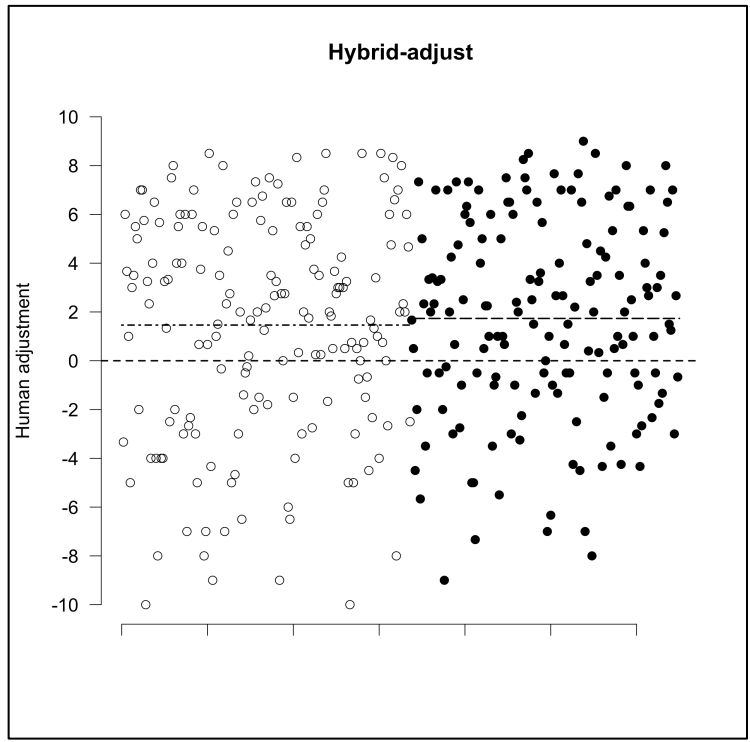

Figure 2. Difference between human judgment and the anchoring position for deceptive (circles) and truthful (dots) answers for the human-adjust condition. Positive values indicate that the judgment was adjusted to be more truthful. Negative values indicate that the judgment was adjusted to the deceptive side.

*Human-machine overlap*

The agreement between the class labels from machine learning and the three human conditions is shown in Table 4. In the human baseline and the hybrid-overrule conditions, the agreement between human and automated decisions was low (< 50%). The high agreement in the hybrid-adjust condition was expected because it did not permit the human judges to depart from the automated decision entirely.

Table 4. *Agreement between automated classification and human decision-making.*

| Automated classification | Human baseline | | Hybrid overrule | | Hybrid adjust | |
|---|---|---|---|---|---|---|
| | Correct | Error | Correct | Error | Correct | Error |
| Correct | 107 (33.13%) | 116 (35.91%) | 101 (32.58%) | 114 (36.77%) | 193 (60.21%) | 28 (7.17%) |
| Error | 55 (17.03%) | 45 (13.93%) | 51 (16.45%) | 44 (14.20%) | 23 (8.72%) | 77 (23.99%) |
| % agreement | 47.06% | | 46.77% | | 84.11% | |

*Partial triaging of uncertain cases*

Lastly, we looked at the human performance of cases that had a machine learning class probability around the 0.50 threshold. A hope attached to hybrid decision systems is that humans can augment the decisions of a machine learning system when the latter is uncertain. Table 5 shows that for three different class probability ranges (+/- 0.02, +/- 0.05, and +/- 0.10), in none of the human conditions does the judgment of humans improve the automated judgment or exceed the random classification performance.

Table 5. *Accuracy for cases in specific class probability range P.*

|  | 0.48 < P < 0.52 | 0.45 < P < 0.55 | 0.40 < P < 0.60 |
|---|---|---|---|
| n | 27 | 101 | 180 |
| Human baseline | 0.52 [0.32; 0.71] | 0.47 [0.37; 0.57] | 0.53 [0.45; 0.60] |
| Hybrid-overrule | 0.60 [0.39; 0.79] | 0.49 [0.39; 0.59] | 0.55 [0.47; 0.62] |
| Hybrid-adjust | 0.50 [0.30; 0.70] | 0.53 [0.43; 0.63] | 0.59 [0.52; 0.67] |
| Machine only | 0.50 [0.30; 0.70]* | 0.56 [0.46; 0.66] | 0.62 [0.55; 0.69] |

Note. Squared brackets denote the 95% confidence interval. * = sign. better than chance level at $p < .001$.

## DISCUSSION

This paper aimed to test whether human judgments can augment the decisions reached with an automated deception detection approach. Using truthful and deceptive statements about people's plans for the next week, an automated machine learning approach achieved a classification performance significantly above the chance level. Although the accuracy reached with the best model on a hold-out test set still implied a considerable error rate (here: 31% errors for a 69% accuracy), this performance similar to a body of research on automated deception detection (Kleinberg et al., 2018; Mihalcea & Strapparava, 2009; Pérez-Rosas & Mihalcea, 2014; Soldner et al., 2019). Human judges, when asked to indicate the likelihood of a statement being deceptive or truthful performed at the chance level. That finding is in line with meta-analytical evidence on human deception detection performance (Bond & DePaulo, 2006; Hartwig & Bond, 2011).

**Human-machine integration**

The central question of this study was whether a combination of machine and human judgments improves the former. Promising findings from other areas indicate that such a combination can indeed improve detection accuracy (Bulten et al., 2020; Jhaver et al., 2019). When human judges were presented with the outcome of the machine learning classification in the current study, the accuracy dropped dramatically from 69% to the chance level when they could freely adjust the prediction. That is, humans impaired the detection accuracy by overruling machine judgment. Specifically, human assessors tended to rate the statements as more truthful than the machine. Since they did so regardless of the actual veracity of the statement, they were able to correctly identify more truth-tellers (76%) than the machine learning approach (60%) but at the cost of a considerably lower lie detection rate (25% vs 76%). The data thus support the truth bias (Levine, 2014, 2014). What our study adds to the body of research supporting the truth bias is that it extends to a setting where humans are supposed to integrate their judgment with that of a machine.

Since full overruling power might give human assessors too difficult a task, we also tested whether constraints on the allowed adjustments enable humans to improve the detection performance. Again, the data suggest that humans acted according to their truth bias and thereby impaired the overall accuracy. It is noteworthy that the truth bias was the most prevalent when humans could fully overrule machine judgment. The tendency to lean towards the truthful default was here driven by the truth bias for deceptive statements which was almost twice as high as that for truthful statements. When human judges had full control, they made deceptive statements more truthful and did so to a much higher degree than for truthful statements. The indiscriminate application of the truth preference meant that the overall accuracy did not exceed the chance level. An interesting aspect is that the same pattern was not found when humans did not receive any information about the machine learning judgment and instead started at a neutral midpoint. In that case, the truth bias was applied to the same degree on truthful and deceptive statements. While research is still scarce on human-machine integration in deception detection, a potential remedy to the incorrect overruling of and possibly mistrust in the machine judgment could come from transparent, explainable machine decisions (Bhatt et al., 2019). The idea of whether humans trust a system more if they understand it is an interesting avenue for future work.

The core finding of this paper is that human involvement in the deception detection process was not beneficial. Instead, machine learning classification resulted in an accuracy in line with typical findings in this research area. Especially for uncertain cases would human involvement have been an interesting addition. However, the current data suggest that the chance performance of humans persists. A possible explanation for the current findings is that the task of deception detection is simply too difficult for humans. Research has repeatedly shown that there is no tell-tale sign like Pinocchio's nose that we can use as a heuristic to determine whether someone is truthful or not. That is not to say that there are no differences between truths and lies, but these are likely small (DePaulo et al., 2003). Consequently, methods that maximize the information we can extract from high-dimensional data should outperform limited human capacity (Hartwig & Bond, 2014). It is then not surprising that machine learning outperforms humans. What the current study shows is that humans not only do not add to an automated detection system, but they actively deteriorate its performance. Our idea that the use of contextual information adds a meaningful layer to context-free, automated decisions is thus not supported.

**Limitations and outlook**

Two limitations are essential to mention. First, the data collection phase asked participants to produce genuine or fabricated statements about their plans for an upcoming week. That setting avoided that liars would need to lie about an event or plan they did not have in the first place. However, since each planned activity is different, it might be that human judges did not have sufficient contextual knowledge to detect implausibility (e.g., they could not know whether a described behaviour is atypical). Extensions of the current approach could address that point by using a setting that is known to all human judges.

Second, the human judges were not told about empirical findings of verbal deception research (e.g., that a lack of detail is often found to be indicative of deception). That decision was deliberate because research on deceptive intentions has not (yet) yielded conclusive results. Future studies could examine whether a triaging system with informed human judges challenges the current findings. Likewise, the reported study did not apply a human-in-the-loop approach that guided the human judge or included trust-building elements (e.g., transparency in the machine decision). Future work could take the human-machine integration further by, for example, highlighting passages that are uncertain or detailing which linguistic indicators the machine learning system deemed particularly deviant.

**Conclusions**

The findings of this paper allow for three conclusions about deception detection accuracy: (1) fully automated machine learning classification performs significantly better than chance; (2) humans' accuracy was indistinguishable from flipping a coin, likely due to the truth bias, and (3) an integration of human and machine judgment did not improve deception detection performance. When humans were allowed to overrule machine judgment, the overall detection performance was drastically impaired. Future research on automated efforts might offer the most promising path forward for deception detection.

**Appendix**

**Appendix 1. *Gender and age information for the sample in the corpus collection phase***

There were no differences between the conditions in gender, $X^2(1) = 0.88$, $p = .363$. Participants in the truthful condition were marginally older ($M$=37.25 years, $SD$=12.31) than those in the deceptive condition ($M$=35.71, SD=11.03), $F(1, 1557) = 6.80$, $p = .009$, although to a negligible effect size (Cohen's $f = 0.07$). There was no difference in the reported motivation to appear convincing, $F(1, 1638) = 0.04$, $p = .827$.

**Appendix 2. *Deviation from the judgment error in each of the human deception detection conditions.***

Table A1. *Deviation from judgment anchor point per condition.*

| Condition | Total deviation (SD) | Deviation truthful (SD) | Deviation deceptive (SD) |
|---|---|---|---|
| Human baseline | 11.99 (20.12) | 12.86 (20.49) | 11.18 (19.79) |
| Hybrid-overrule | 16.10 (25.21) | 11.00 (24.29) | 20.76 (25.19) |
| Hybrid-adjust | 1.59 (4.34) | 1.73 (4.11) | 1.46 (4.56) |

Note. Positive values indicate that the judgment was adjusted to be more truthful. Negative values indicate that the judgment was adjusted to the deceptive side.